\documentclass[runningheads]{llncs}
\usepackage{graphicx}

\usepackage{tikz}
\usepackage{comment}
\usepackage{amsmath,amssymb} 
\usepackage{color}

\usepackage{array}
\usepackage{multirow}
\usepackage{bm}
\usepackage{caption}
\usepackage{subcaption}
\usepackage{floatrow}

\usepackage[tableposition=top]{caption}
\usepackage{float}
\floatstyle{plaintop}
\restylefloat{table}
\newfloatcommand{capbtabbox}{table}

\begin{document}
\pagestyle{headings}
\mainmatter
\def\ECCVSubNumber{5648}  

\title{Head2HeadFS: Video-based Head Reenactment with Few-shot Learning} 

%
\author{Michail Christos Doukas\inst{1} \and Mohammad Rami Koujan\inst{2} \and
Viktoriia Sharmanska\inst{1} \and
Stefanos Zafeiriou\inst{1}}
\authorrunning{M. C. Doukas et al.}
%
\institute{$^1$ Department of Computing, Imperial College London, UK \\
$^2$ College of Engineering, Mathematics and Physical Sciences, University of Exeter, UK}
\maketitle

\begin{abstract}
Over the past years, a substantial amount of work has been done on the problem of facial reenactment, with the solutions coming mainly from the graphics community. Head reenactment is an even more challenging task, which aims at transferring not only the facial expression, but also the entire head pose from a source person to a target. Current approaches either train person-specific systems, or use facial landmarks to model human heads, a representation that might transfer unwanted identity attributes from the source to the target. We propose head2headFS, a novel easily adaptable pipeline for head reenactment. We condition synthesis of the target person on dense 3D face shape information from the source, which enables high quality expression and pose transfer. Our video-based rendering network is fine-tuned under a few-shot learning strategy, using only a few samples. This allows for fast adaptation of a generic generator trained on a multiple-person dataset, into a person-specific one.
\end{abstract}

\section{Introduction}

The task of facial reenactment \cite{Liu2001}, \cite{Olszewski2017} , \cite{Suwajanakorn2017}, \cite{thies2015realtime}, \cite{face2face}, \cite{Vlasic2005} aims at transferring facial expressions from a driving (source) person to another (target) person. In most cases, this is done by modifying the deformations solely within the inner facial region of the target actor. Thus, even when expression transfer is successful, the final reenactment result often looks non-plausible due to misalignment of the expressions and the head movements. 
In this paper, we address the task of head reenactment, 
where the objective is transferring the entire head pose and movement of the source person to the target person, including but not limited to the facial expressions. This is a particularly challenging problem, with applications that range from video editing and movie dubbing to telepresence and virtual reality. Recent works have been able to produce impressive results \cite{head2head}, \cite{head2headpp}, \cite{deepvideoportraits}, \cite{headon}, \cite{fewshot}. Nonetheless, many of these attempts come with limitations in terms of transferability performance, photo-realism, as well as training and test execution time.

Warping-based approaches on head reenactment, such as \cite{elor2017bringingPortraits}, \cite{X2Face} produce quite promising results, although they are not always reliable in terms of realism. On the other hand, computer graphics based methods  \cite{headon} create highly realistic videos, nonetheless, they usually operate on datasets collected under constrained conditions and special capturing tools, such as RGB-D cameras. Approaches such as \cite{deepvideoportraits}, which capitalize on 3DMMs \cite{blanz1999morphable} and deep learning architectures, require training a separate model for each target person, which is especially time consuming, as it requires many hours of training for each individual. Very recent methods \cite{fewshot} \cite{vid2vidfs} tackle this issue with few-shot learning. However, these approaches use facial landmarks as a pose and expression representation, which creates other issues, such as transferring also identity attributes from the source to the target person.

Our proposed \textit{head2headFS} method overcomes the aforementioned limitations, as it utilizes both a 3D facial reconstruction stage and few-shot learning techniques. More specifically, our pipeline consists of a robust 3D face reconstruction and tracking system, and a video rendering neural network. Owing to the few-shot learning ability, our suggested framework can learn to reenact a target person from only a few tens of frames, reducing the training time considerably compared to person-specific approaches and, consequently, making it more suitable for real-world applications. We condition frame synthesis on \textit{Normalised Mean Face Coordinates (NMFC)} images, a geometry-based head representation obtained from the 3D facial reconstruction stage . This representation allows transferring the pose and expression from any source to a given target person, without distorting the identity of the latter. Our video rendering network is trained initially on a large dataset of 500 different identities. After that, it can be quickly fine-tuned on a new target person, using a short training video of 50-500 frames, in a few-shot manner. During test, our end-to-end pipeline performs head reenactment from a source to the selected target in real-time.

\section{Related Work}

\setlength\parindent{0pt} 
\setlength{\parskip}{2pt}

\textbf{3D Face Reconstruction.} The problem of recovering the 3D geometry of human faces from monocular images has been recently targeted considerably due to its centrality in many applications, ranging from facial reenactment, recognition, human-computer interactions, performance capture and tracking, etc. Due to their pose and illumination invariance, 3D facial data constitute a dispensable geometrical description of
faces for various facial image processing systems. The field itself is rich in approaches targeting this problem under different assumptions and constraints. One line of research relies on 3D Morphable Models (3DMMs) for approaching this problem. 3DMMs have been used substantially in the literature since the pioneering work of Blanz and Vetter \cite{blanz1999morphable}, with many extensions \cite{koujan2018combining}, \cite{booth20183d}, \cite{faggian20083d}, \cite{huber2016multiresolution}, \cite{li2017learning}. They are generative parametric models built from a set of 3D facial scans, coupled to each other with anatomical correspondences, and can represent any unseen faces as a linear combination of the training set. We follow this line of research and utilise 3DMMs in this work for performing the 3D reconstruction from videos.

\textbf{Facial and Head Reenactment Methods.} Much research has been focused on the problem of facial reenactment \cite{Olszewski2017}, \cite{thies2015realtime}, \cite{face2face}. Many works transfer expression either with 2D warping techniques \cite{Liu2001}, \cite{Garrido2014}, or by utilizing 3D face models \cite{Vlasic2005}, \cite{thies2015realtime}, \cite{face2face}, \cite{DeferredNeuralRendering}. These methods do not provide complete control over the generated video, as they manipulate only the interior of the face. Recently, there has been a growing amount of attempts in the direction of both expression and pose transfer \cite{headon}, \cite{elor2017bringingPortraits}, \cite{X2Face}, \cite{fewshot}, \cite{deepvideoportraits}. One of the first approaches, X2face \cite{X2Face}, follows a warping-based approach that causes deformations in the generated heads. Zakharov \emph{et al.} \cite{fewshot} propose a few-shot, image-based adversarial learning approach, as their network learns to generate unseen target identities even from a single image. Nonetheless, their neural network relies on landmarks, which leads to identity distortions of the target person, since facial landmarks include identity related information from the source. As far as we are aware, Deep Video Portraits \cite{deepvideoportraits} is the only learning-based head reenactment system that uses 3D facial information to condition video synthesis. However, this model requires a long video footage of the target person and training a new model for each target takes many hours.

\textbf{Video Synthesis.} Generative Adversarial Networks (GANs) \cite{ganGoodfellow} have been extensively used for image or video synthesis. Synthesis is usually conditioned either on noise distributions \cite{ganGoodfellow}, \cite{PGAN}, or other data types. For instance, there is a large body of work on image-to-image translation \cite{UNIT}, \cite{TaigmanPW16}, \cite{pix2pix}, \cite{pix2pixHD} and conditional video synthesis \cite{vid2vid}, \cite{Saito2017TemporalGA}, \cite{VondrickNIPS2016}, \cite{tulyakov2017mocogan}, \cite{Denton2017}, \cite{Finn2016}, \cite{Chen_2017_ICCV}. In our work, we adopt a video-based GAN framework, which renders photo-realistic and temporally coherent frames.

\section{Methodology}



\setlength\parindent{12pt} 
\setlength{\parskip}{0pt}

Our \textit{head2headFS} pipeline is composed of two sequential stages: a) 3D facial reconstruction and tracking, and b) deep video rendering network. Our deep video rendering stage capitalises on a carefully designed GAN-based model. We train such a model in two subsequent stages: 1) an initialization stage, where the generative network learns to synthesize multiple people and 2) a fine-tuning stage, in which our model is fine-tuned in a few-shot setting to generate a new target person.

\subsection{3D Facial Recovery}
Our goal here is to generate a reliable estimation of the human facial geometry, capturing the temporal dynamics while being separable in each frame into the identity and expression contributions of the videoed subject. We relay on such a separability to effectively disentangle the human head characteristics in a transferable and photo-realistic way between different facial videos. Towards that aim, we benefit from the prior knowledge in our problem space and  harness the power of 3DMMs for 3D reconstructing and tracking the faces appearing in the input sequences. Given a sequence of $T$ frames $\mathcal{F}_{1:T}= \{f_t \mid {t=1,\dots,T}\}$, the 3D reconstruction and tracking stage produces two sets of parameters: 1) shape parameters $\mathcal{S}=\{\bm{s_t} \mid \bm{s_t} \in \mathbb{R}^{n_i+n_e}, t=1,...,T\}$ , and 2) camera parameters $\mathcal{P}=\{\bm{p_t} \mid \bm{p_t} \in \mathbb{R}^{7}, t=1,...,T\}$, embodying rotation, translation and orthographic scale.

\setlength\parindent{0pt} 
\setlength{\parskip}{2pt}

\textbf{Shape representation.} Under 3D Morphable Models, a 3D facial shape $\mathbf{x}_t=[x_1, y_1, z_1,..., x_N, y_N, z_N]^T \in \mathbb{R}^{3N}$ can be represented mathematically as:
\begin{equation}
    \label{eq:3DMM}
\mathbf{x}(\bm{s_t}^i,\bm{s_t}^e)=\bar{\mathbf{x}}+\mathbf{U}_{id} \bm{s_t}^i+ \mathbf{U}_{exp} \bm{s_t}^e
\end{equation}
where $\mathbf{\bar{x}}\in \mathbb{R}^{3N}$ is the overall mean shape vector of the morphable model, given by $\mathbf{\bar{x}}=\mathbf{\bar{x}}_{id}+\mathbf{\bar{x}}_{exp}$,  with $\mathbf{\bar{x}}_{id}$ and $\mathbf{\bar{x}}_{exp}$ depicting the mean identity and expression of the model, respectively. $\mathbf{U}_{id} \in \mathbb{R}^{3N\times n_i}$ is the identity orthonormal basis with $n_i$ principal components ($n_i\ll3N$), $\mathbf{U}_{exp} \in \mathbb{R}^{3N\times n_e}$ is the expression orthonormal basis with the $n_e$ principal components ($n_e\ll3N$) and $\bm{s_t}^i \in \mathbb{R}^{n_i}$, $\bm{s_t}^e \in \mathbb{R}^{n_e}$ are the identity and expression parameters of the morphable model. We designate the joint identity and expression parameters of a frame $\bm{t}$ by $\bm{s_t}=(\bm{s_t}^{i}, \bm{s_t}^e)$.  In the adopted model \eqref{eq:3DMM}, the 3D facial shape $\mathbf{x}$ is a function of both identity and expression coefficients ($\mathbf{x}(\bm{s_t}^i, \bm{s_t}^e)$), with expression variations being effectively represented as offsets from a given identity shape.

\textbf{3D reconstruction from videos.} We adopt here a fast and robust sparse landmark-based method that capitalises on the rich temporal information in videos while performing the reconstruction. We rely on the fact that state-of-the-art facial alignment methods are quite robust and accurate, and use the method of \cite{guo2018stacked} to extract 68 landmarks from each frame.  While carrying out the 3D reconstruction,  we postulate scaled orthographic projection (SOP) and assume that in each video the identity parameters $\bm{s_t}^i$ are fixed (yet unknown) throughout the entire video, letting however the expression parameters $\bm{s_t}^e$ as well as the camera parameters (scale and 3D pose) to differ among frames.
In sum, for a given sequence of frames, we minimise an energy equation that consists of three terms as demonstrated in equation \eqref{eq: energy_eq} : \textbf{a)} a data term penalising the $\ell_{2}$ norm error of the projected landmarks over all frames ($E_{l}$),  \textbf{b)} shape regularisation term ($E_{pr}$) that reinforces a quadratic prior over the identity and per-frame expression parameters and \textbf{c)} a temporal smoothness term ($E_{sm}$) that supports the smoothness of the expression parameters throughout the video, by employing a quadratic penalty on the second temporal derivatives of the expression vector.
\begin{equation}
    \label{eq: energy_eq}
    E(\mathcal{S}, \mathcal{P})= w_{l} E_{l}(\mathcal{S}, \mathcal{P}) + w_{pr} E_{pr}(\mathcal{S}) + w_{sm} E_{sm}(\mathcal{S}^e)
\end{equation}

\setlength\parindent{12pt} 
\setlength{\parskip}{0pt}

We also tackle the problem of gross occlusions by imposing box constraints on the identity and per-frame expression parameters. After estimating the camera parameters ($\mathcal{P}$) in \eqref{eq: energy_eq} in an initialisation stage, the minimisation of the loss function leads to a large-scale least squares problem with box constraints, which we solve efficiently by using the reflective Newton method of \cite{coleman1996reflective}. We refer the reader to the supplementary materials for more details about the initialisation stage of our 3D reconstruction and tracking step.

\setlength\parindent{0pt} 
\setlength{\parskip}{2pt}

\textbf{Adopted 3DMM characteristics.} For all the experiments conducted in this work, we adopt a large-scale 3DMM provided by the authors of \cite{booth20163d}, \cite{booth2018large} for the identity part $\{\mathbf{\bar{s}}_{id}, \mathbf{U_{id}}\}$. Their 3DMM was constructed from approximately 10,000 scans of different people, making it the biggest 3DMM ever constructed, with varied demographic information. In addition, the expression part of the model, $\{\mathbf{\bar{x}}_{exp}, \mathbf{U}_{exp}\}$ originates from the work of Zafeiriou et al.~\cite{Zafeiriou2017}, who built it using the blendshapes model of Facewarehouse \cite{cao2014facewarehouse} with the help of Nonrigid ICP \cite{cheng2017statistical} to register the blendshapes model with the LSFM model.

\subsection{Facial Conditioning Images}

\setlength\parindent{12pt} 
\setlength{\parskip}{0pt}

Our video renderer expects as input:
1) an RGB frame sequence of the target person and 2) the facial parameterisation of each frame, extracted from the given sequence by the face reconstruction and tracking step. Such a parameterisation disentagles identity from expression, allowing us to train our video rendering network on a specific target person, and transfer the expression and pose of another source subject with different head characteristics, in the test phase. To aid our neural renderer in learning this mapping accurately, we transfer the head parameterisation we extract during the 3D reconstruction step to a more representative space. More specifically, given the estimated head parameters of a frame $t$ (shape $\bm{s_t}$ and camera $\bm{p_t}$), we rasterize the 3D reconstruction of this frame, producing a visibility mask ($\mathbf{M} \in \mathbb{R}^{W\times H}$) in the image space. Each pixel of $\mathbf{M}$ stores the ID of the corresponding visible triangle on the 3D face from this pixel. We thereafter store the normalised x-y-z coordinates, coming from the mean face of the utilised 3DMM, of the centre of this triangle  in another image, we call in this paper the Normalised Mean Face Coordinates ($\mathbf{NMFC}) \in \mathbb{R}^{W\times H \times3}$ image, and utilise it as the conditional input of the video rendering stage. Equation (\ref{eq: NMFC}) detials this process.
\begin{equation}
    \label{eq: NMFC}
    \mathbf{NMFC}_t= \mathcal{E}(\mathcal{R}(\bm{x_t}(\bm{s_t}^i, \bm{s_t}^e), \bm{p_t}), \bar{\bm{x}}),
\end{equation}
where $\mathcal{R}$ is the rasterizer, $\mathcal{E}$ is the encoding function and $\bar{\bm{x}}$ is the normalised version of the utilised 3DMM mean face (see (\ref{eq:3DMM})), so that the x-y-z coordinates of this face $\in [0, 1]$. This representation is more compact and easy-to-interpret by our video rendering network since it associates well with the corresponding RGB frame, pixel by pixel, and, subsequently, leads to a realistic and novel video synthesis. Likewise, during the test time, we similarly create the $\mathbf{NMFC}$ images but with $\bm{s_t}^e$ and $\bm{p_t}$ representing the facial expression and pose of the source person at frame $t$, while the identity coefficients $\bm{s_t}^i$ are taken from the target. Even though this geometry-based representation ignores some facial parts, inner-mouth region and hair, those can be well-captured by the video rendering stage during the fine-tuning phase, see section \ref{sec:experiments}.



\begin{figure*}[h!]
    \centering
    \includegraphics[width=\linewidth]{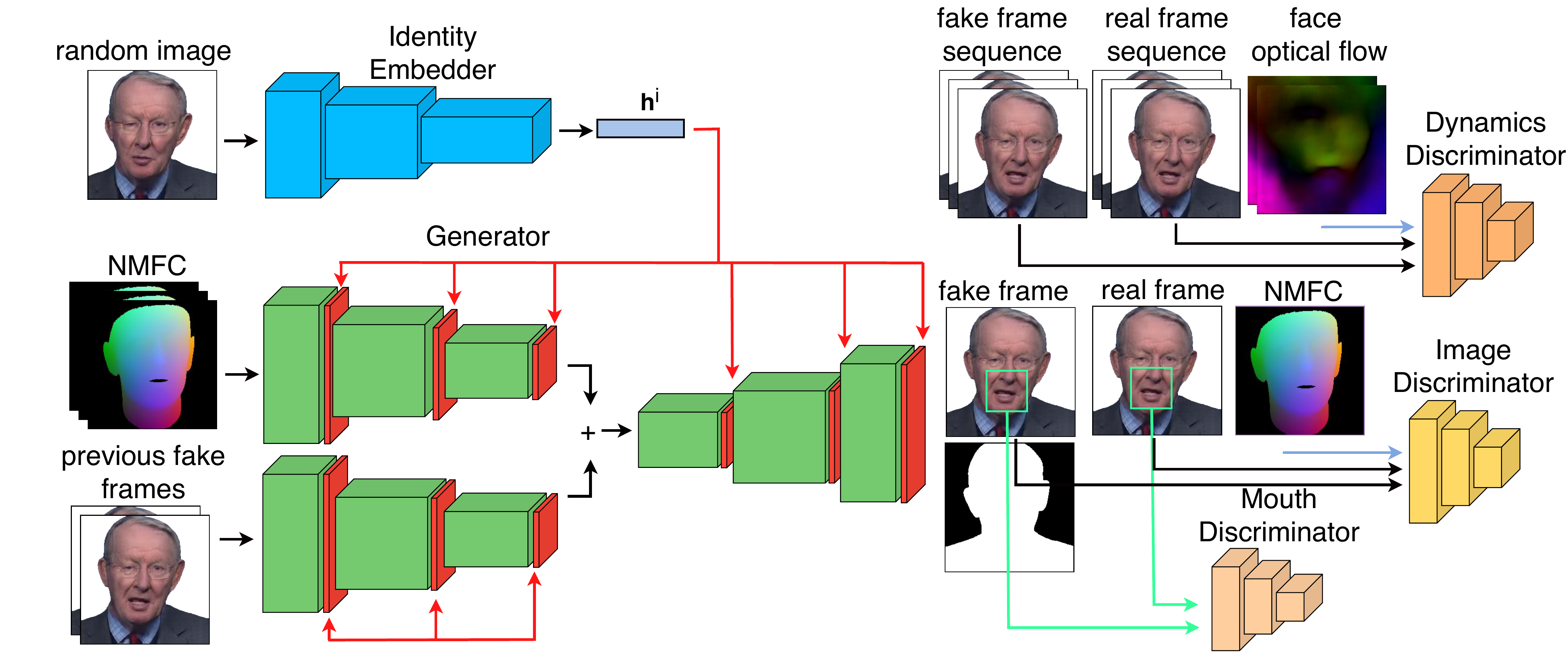}
    \caption{Our framework during the initialization training stage. Synthesis is conditioned both on NMFC images and frames generated in previous time steps.
    The identity feature vector $\bm{h^i}$ computed by the identity embedder is injected to the generator through the adaptive instance normalization layers.
    }
    \label{fig:fig1}
\end{figure*}

\subsection{Video Rendering Network Initialization Stage}

\setlength\parindent{12pt} 
\setlength{\parskip}{0pt}

In this first learning phase of our video rendering network, we utilize a multi-person dataset with $N$ identities. Let $\bm{Y}_{1:T}^i = \{\bm{y}_t^i \}_{t=1,\dots,T}$ be a video of the $i$-th target person in our training corpus, with removed background, and $\bm{M}_{1:T}^i = \{\bm{m}_t^i \}_{t=1,\dots,T}$ the corresponding foreground masks, extracted from the original video. We apply our facial reconstruction method to obtain the corresponding NMFC frame sequence $\bm{X}_{1:T}^i = \{\bm{x}_t^i \}_{t=1,\dots,T}$, for each identity $i=1, \dots, N$. 
Then, our generative network $G$ aims to learn a mapping from this conditioning representation to a photo-realistic and temporally coherent video $\bm{\tilde{Y}}_{1:T}^i$, as well as a prediction of the foreground mask $\bm{\tilde{M}}_{1:T}^i$. This is done in a self-reenactment setting, thus the generated video should be a reconstruction of $\bm{Y}_{1:T}^i$, which serves as a ground truth.

Apart from face expression and head pose, we further condition synthesis on identity details of the target person $i$. Despite the fact that the NMFC sequence contains identity related information coming from the 3DMM, such as the head shape and geometry, it does not provide any knowledge on skin texture, hair, or upper body details. We totally remove the need to learn the identity from the NMFC input, by incorporating an identity embedding network $E_{id}$ to our framework. This network learns to compute an identity feature vector $\bm{h}^i \in \mathbb{R}^{n_f}$, which is passed to the generator to condition synthesis. Our proposed framework additionally consists of an image discriminator $D^I$, a dynamics discriminator $D^D$ and a dedicated mouth discriminator $D^M$, which ensure that the generated video looks realistic and temporally coherent at the same time. In the following, we explain the role and significance of each component.

\setlength\parindent{0pt} 
\setlength{\parskip}{2pt}

\textbf{Identity Embedder $E_{id}$.} Given a target person $i$, we randomly select $M$ frames $\bm{y}_{m_1}^i, \dots, \bm{y}_{m_M}^i$ from the training video $\bm{Y}_{1:T}^i$. We pass each one of them to the embedder and then compute the identity feature vector:
\begin{equation}
    \label{eq: Eid}
    \bm{h}^i = \dfrac{1}{M} \sum_{j=1}^M E_{id} (\bm{y}_{m_j}^i;\theta_{E_{id}})
\end{equation}
Our choice to pick random frames from the training sequence $i$ and average the embedding vectors, automatically renders $\bm{h}^i$ independent from the person's head pose appearing in the $M$ random input images.

\textbf{Generator $G$.} Our novel generator conditions on two separate information sources: a) $\bm{X}_{1:T}^i$, the head pose and expression representation (NMFC) and b)
$\bm{h}^i$, the identity representation of the target person. We employ a sequential generator, which produces frames and masks one after the other, similarly with \cite{vid2vid}. We model the dependence of the synthesized frame $\bm{\tilde{y}}_t^i$ and the predicted foreground mask $\tilde{\bm{m}}_{t}^i$ on past time steps by conditioning synthesis on the two previously generated frames $\bm{\tilde{Y}}_{t-2:t-1}^i = \{\bm{\tilde{y}}_{t-2}^i, \bm{\tilde{y}}_{t-1}^i \}$. The synthetic frame and the hallucinated foreground mask at $t$ are computed by the generator, as
\begin{equation}
    \label{eq: G}
    \tilde{\bm{y}}_{t}^i, \tilde{\bm{m}}_{t}^i = G(\bm{X}_{t-2:t}^i, \bm{\tilde{Y}}_{t-2:t-1}^i, \bm{h}^i_*; \theta_G),
\end{equation}
where $\bm{X}_{t-2:t}^i = \{\bm{x}_{t-2}^i, \bm{x}_{t-1}^i, \bm{x}_{t}^i \}$ is the conditional input, corresponding to the current step and the two previous ones, while $\bm{h}^i_*$ is the final identity feature vector, which is kept fixed for the synthesis of the entire sequence.

\setlength\parindent{12pt} 
\setlength{\parskip}{0pt}

The generator consists of two identical encoding blocks and one decoding block. We feed the conditional input $\bm{X}_{t-2:t}^i$ to the first encoder and the previously generated frames $\bm{\tilde{Y}}_{t-2:t-1}^i$ to the second one. Then, their resulting features are summated and passed to a decoder, which outputs the generated frame and mask $\tilde{\bm{y}}_{t}^i$, $\tilde{\bm{m}}_{t}^i$. The identity feature vector
$\bm{h}^i$ is injected into the generator through its normalization layers. We employ a layer normalisation method based on AdaIN \cite{adain}, similar to the one presented in \cite{fewshot}. However, we apply this method to all normalisation layers, both in the encoding and decoding blocks of $G$. More specifically, given a normalization layer in $G$ with input $x$, the corresponding modulated output is calculated as
\begin{equation}
    \label{eq: AdaIN}
    N(x, \bm{h}^i) = (\bm{P}_{\gamma} \bm{h}^i) \dfrac{x - \mu(x)}{\sigma(x)} + (\bm{P}_{\beta} \bm{h}^i),
\end{equation}
where $\bm{P}_{\gamma}$ and $\bm{P}_{\beta}$ matrices are the learnable parameters, which project the  identity feature vector to the affine parameters of the normalization layer.

\setlength\parindent{0pt} 
\setlength{\parskip}{2pt}

\textbf{Spatial discriminators $D^I$ and $D^M$.} The principal role of these networks is to increase the photo-realism of generated frames, as they learn to distinguish real from fake images. Given a uniformly sampled time step $t \in  [0, T]$, the convolutional part of the image discriminator $D^I$ receives the real $\bm{y_t^i}$, or the synthesized frame $\bm{\tilde{y}_t^i}$, along with the corresponding conditional input $\bm{x}_t^i$, and computes a feature vector $\bm{d}$. Moreover, similar to \cite{fewshot}, our image discriminator keeps a learnable matrix $\bm{W} \in \mathbb{R}^{N \times n_f}$, with each one of its lines representing a different person in the dataset. Given the identity index $i$, the image discriminator chooses the appropriate row $\bm{w}_i$. Then, the realism score for person $i$ is calculated as:
\begin{equation}
    \label{eq: DI}
    r = \bm{d}^{\top} (\bm{w}_i + \bm{w}_0) + c
\end{equation}
where $\bm{w}_0$ and $c$ are speaker independent learnable parameters of $D^I$. In this way, $r$ reflects whether or not the head in the input frame is real and belongs to identity $i$ and at the same time corresponds to the NMFC conditional input $\bm{x}_t^i$. By selecting each time the appropriate row vector of $\bm{W}$, given the identity index $i$, the vectors $\bm{w_i}$ acquire person specific values during training, resembling identity embeddings. Apart from the image discriminator, a mouth discriminator $D^M$ is employed to further improve the visual quality of the mouth area, since teeth synthesis is a particularly challenging task. This network receives the cropped mouth regions from the real frame $\bm{y_t^i}$ or the fake one $\bm{\tilde{y}_t^i}$ and computes the realism score of the mouth area.

\textbf{Temporal discriminator $D^V$.} In spite the fact that the image and mouth discriminators classify individual frames as being real or fake, they do not provide any feedback to $G$ regading the temporal dynamics of the video. To that end we introduce $D^V$, a dynamics discriminator which learns to distinguish between realistic and non-realistic temporal dynamics in videos. A sequence of $K$ consecutive frames $\bm{Y}_{t:t+K-1}^i$ is randomly drawn from the ground truth video, along with $K$ frames $\bm{\tilde{Y}}_{t:t+K-1}^i$ from the generated one. Apart from the real or fake sequences, $D^V$ also observes the optical flow $\bm{V}_{t:t+K-2}$ extracted from $\bm{Y}_{t:t+K-1}^i$. Therefore, the realism score reflects on whether or not the optical flow agrees with the motion in the short input video, forcing the generator to synthesize temporally coherent frames.

\textbf{Objective functions.} All the components of our framework are jointly trained under an adversarial setting. The parameters of both the video generator and identity embedder are optimized under the adversarial loss $\mathcal{L}_{adv} = \mathcal{L}_{adv}^{D^I} + \mathcal{L}_{adv}^{D^M} + \mathcal{L}_{adv}^{D^V}$, with each loss term coming from the corresponding discriminator network. As proposed in \cite{fewshot}, we add an embedding matching loss to the objective of $E_{id}$, taking advantage of the identity representation $\bm{W}$, learned by the image discriminator, for each identity in the dataset. More specifically, given the person $i$, the identity feature vector $\bm{h}^i$ computed by the identity embedder and the corresponding row $\bm{w}_i$ of matrix $\bm{W}$, we calculate the cosine distance between the identity features:
\begin{equation}
    \label{eq: MCH}
    \mathcal{L}_{mch} = 1 - \dfrac{{\bm{h}^i}^{\top} \bm{w}_i}{||\bm{h}^i||_2 ||\bm{w}_i||_2}.
\end{equation}

\setlength\parindent{12pt} 
\setlength{\parskip}{0pt}

Then, the total objective of the identity embedder becomes $\mathcal{L}_{E_{id}} = \mathcal{L}_{adv} + \lambda_{mch} \mathcal{L}_{mch}$, which is minimized. Moreover, we add three more terms in the loss function of the generator $G$, a VGG loss, a feature matching loss \cite{pix2pixHD}, \cite{vid2vid} and a mask reconstruction loss. Given a ground truth frame $\bm{y}^i_t$ and the synthesised frame $\tilde{\bm{y}}^i_t$, we use the pre-trained VGG network \cite{vgg} to compute the VGG loss $\mathcal{L}_{vgg}$.
The feature matching loss $\mathcal{L}_{feat}$ is calculated by extracting features with the two discriminators $D_I$ and $D_V$ and computing the $\ell_1$ distance between the features extracted from the fake frame $\tilde{\bm{y}}^i_t$ and the corresponding ground truth $\bm{y}^i_t$. For the mask reconstruction loss, we compute the simple $\ell_1$ distance between the ground truth foreground mask and the one predicted by $G$.
The total objective of $G$ can be written as $\mathcal{L}_G = \mathcal{L}_{adv} + \lambda_{vgg} \mathcal{L}_{vgg} + \lambda_{feat}\mathcal{L}_{feat} + \lambda_{mask}\mathcal{L}_{mask}$. Finally, all discriminators are optimised alongside $E_{id}$ and $G$, under the corresponding adversarial objective functions. An overview of our framework during the initialization stage is shown in Fig. \ref{fig:fig1}.

\subsection{Video Rendering Network Fine-tuning Stage}

In this second training stage, we fine-tune the networks of our framework to a new unseen identity, by taking advantage of the network parameters learned from the multiple person dataset, in the previous training stage. In this way, we are able to obtain a strong person specific generator very quickly, using a very small number of training samples, in just a few epochs. Given the new short frame sequence $\bm{Y}_{1:T'}^{new}$ and the extracted foreground masks $\bm{M}_{1:T'}^{new}$, first we use our facial reconstruction model to compute the generator's conditional input $\bm{X}_{1:T'}^{new}$. Then, we pass each RGB frame through the identity embedder and calculate the average identity feature
\begin{equation}
    \label{eq: avgid}
    \bm{h}^{new} = \dfrac{1}{T'} \sum_{j=1}^{T'} E_{id} (\bm{y}_{j}^{new}),
\end{equation}
which serves as a representation of the new target person.
After computing $\bm{h}^{new}$, the embedder $E_{id}$ is no longer needed in the fine-tuning stage.

\setlength\parindent{0pt} 
\setlength{\parskip}{2pt}

\textbf{Generator initialization.} This vector is used to initialize the normalization layers of $G$. We replace each AdaIN normalization layer of $G$, with a simple instance normalization layer \cite{Ulyanov2016InstanceNT}. Then, the identity projection matrices $\bm{P}_{\gamma}$ and $\bm{P}_{\beta}$, learned during the initialization stage, are multiplied with $\bm{h}^{new}$ and the resulting vectors
$\gamma = \bm{P}_{\gamma} \bm{h}^{new}$ and $\beta = \bm{P}_{\beta} \bm{h}^{new}$
are used as an initialization of the  modulation parameters of the instance normalization layer. The rest parameters of $G$ are simply initialized from the values learned in the first multi-person training stage.

\textbf{Discriminators initialization.} Since we no longer present images of different people to $D^I$, we replace the matrix $\bm{W}$ that contains an identity representation vector for each person in the multi-person dataset, with a single vector $\bm{w}$, which plays the role of row $\bm{w}_i$ and is initialized with the values of $\bm{h}^{new}$. The convolutional part of $D^I$ is initialized with the values learned from the previous training stage, which also happens for the other two discriminators $D^M$ and $D^V$.

\textbf{Training.} After setting the initial values for  the generator and the discriminators from the previous stage, the framework is trained in an adversarial manner, with the generator aiming to learn the mapping from the NMFC sequence $\bm{X}_{1:T'}^{new}$ to the RGB video $\bm{Y}_{1:T'}^{new}$ and foreground mask $\bm{M}_{1:T'}^{new}$.

\textbf{Synthesis.} Despite the fact that the generator is trained in a self-reenactment setting,  it can be used to perform source-to-target expression and pose transfer during test time. Given a sequence of frames from the source, first we perform 3D reconstruction and compute the NMFC images for each time step, by adapting the reconstructed head shape to the identity parameters of the target person. These NMFC frames are fed to $G$, which synthesizes the desired photo-realistic frames one after the other.

\section{Experiments}
\label{sec:experiments}
\textbf{Dataset.} Our database consists of 500 different identities. We collected around 1000 frames for each person, from publicly-available videos. Out of the 500 identities, we used 480 for training our model in the initialization stage and the rest 20 for fine-tuning a new model for each target person. All videos have a $256 \times 256$ resolution and 20 fps displaying rate. We split each frame sequence in a training and a test part, with last 100 frames being kept for test. During fine-tuning, we randomly select a subset of 50 to 500 subsequent frames from the training split of the target person. We show that even with 50 training frames, our method learns highly realistic heads.

\setlength\parindent{12pt} 
\setlength{\parskip}{0pt}

As a pre-processing step, we used a modification of \cite{Yu_2018_ECCV} \footnote{https://github.com/zllrunning/face-parsing.PyTorch$\#$Demo} to compute a foreground mask and remove the background for each frame. This helps the identity embedder to focus on identity related features, without the interference of noisy and dynamic backgrounds. Then, since our generator learns to predict the foreground mask, during test time we are able to place the background back to the synthesized video or chose a new one.

\setlength\parindent{0pt} 
\setlength{\parskip}{2pt}

\textbf{Implementation details.} We base our generator and discriminator networks architecture on the state-of-the-art video to video translation framework, vid2vid \cite{vid2vid}, For the identity embedder we use the architecture of facenet \cite{facenet} \footnote{https://github.com/timesler/facenet-pytorch}, with a feature vector size $n_f = 512$. Moreover, in practice we employ a multiple scale temporal discriminator, which operates in three different temporal scales. The first scale receives frame sequences in the original frame rate.  Then, the two extra scales are formed by sub-sampling the frames by a factor of two for each scale. In order to force the generator to produce finer details in local patches of the frames, we use PatchGAN  \cite{pix2pix}, \cite{vid2vid}, both for the image and mouth discriminators. For the extraction of optical flow, we use FlowNet2 \cite{flownet2} and we fine-tune it on the 4DFAB dataset \cite{cheng20184dfab}, which contains dynamic 4D  videos of heads. To create the ground truth 2D flow, we use the provided camera parameters and rasterize the 3D scans of 750K frames. We use a hinge loss \cite{sagan} variation for the adversarial objectives of the generator and the discriminators, the same one with \cite{fewshot}. We set all the hyperparameters $\lambda_{mch} = \lambda_{vgg} = \lambda_{feat} = \lambda_{mask} = 10$. Networks are optimized with Adam \cite{adam}, for 20 epochs in the first initialization training stage and 15 epochs in the fine-tuning stage, on a single NVIDIA GeForce RTX 2080 Ti. Fine-tuning a new model on an unseen person, requires as little as 3-30 minutes, depending on the size of the training sequence (50-500 frames). During test time, our entire pipeline can perform \textit{head reenactment from a source person to a target one in real-time} (20 fps).

\setlength\parindent{0pt} 
\setlength{\parskip}{2pt}



\begin{figure}[h!]
    \centering
    \includegraphics[width=\linewidth]{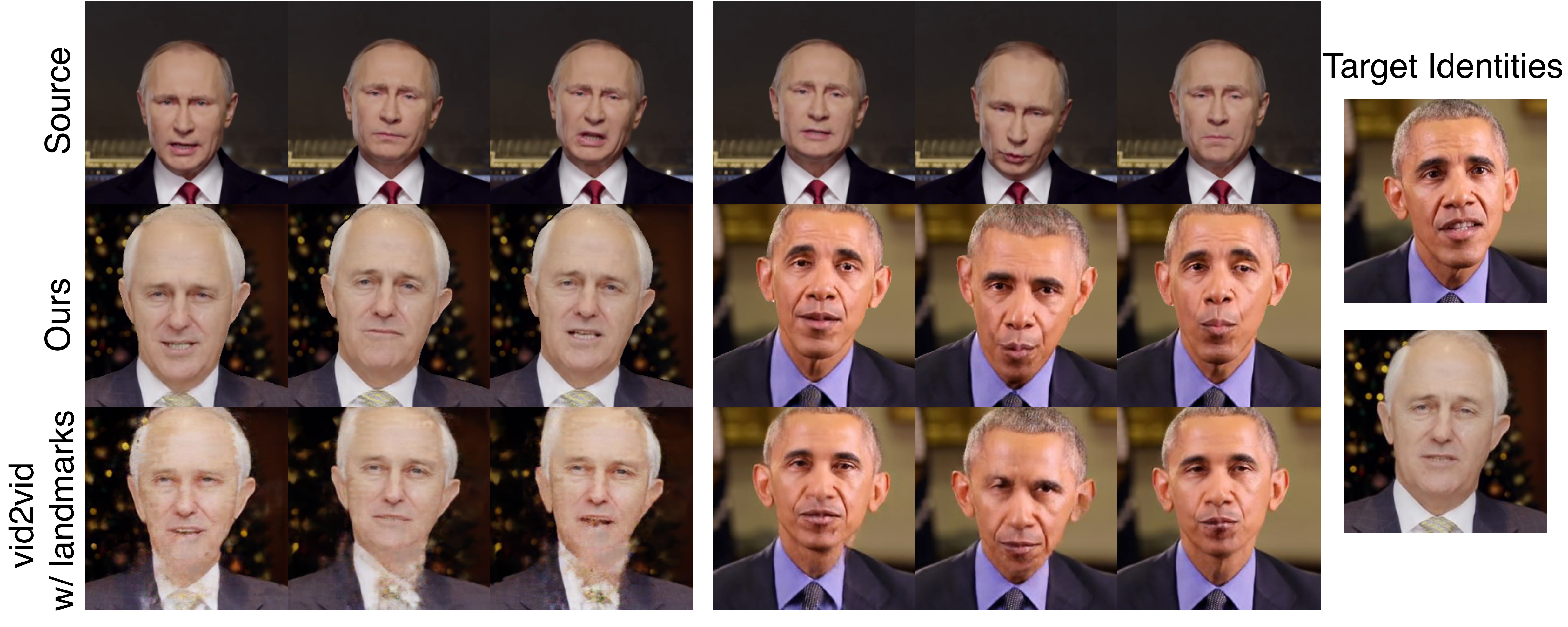}
    \caption{Head reenactment results with vid2vid \cite{vid2vid} conditioned on landmarks. The facial geometry of the source is reflected on the head shape of target.}
    \label{fig:fig2}
\end{figure}

\textbf{Importance of 3D face modeling.} We condition synthesis on 3D facial information, since landmark-based methods are not suitable for arbitrary source to target head reenactment. To demonstrate the significance of the NMFC representation, we conducted a head reenactment experiment conditioning on landmarks. To that end, we trained two person-specific vid2vid models on the task of landmark-to-RGB video translation, using 5K frames from each target identity. During test, we performed landmark size adaptation, from the head size of the source to the head size of the target. We demonstrate the results in Fig. \ref{fig:fig2}. As can be seen, in both cases, the facial shape of the source has been transferred along with the pose and expression.

\textbf{Evaluation Metrics.} For the qualitative evaluation of our models and comparison with other methods, we perform tests under the self-reenactment setting. Given a new identity, we fine-tune a new model on a part of its training set and then use the last 100 left-out frames for test. This gives us access to the ground truth frames, which are then compared to the synthesized ones. We evaluate the quality of the reconstruction, thus the performance of our generator using two metrics: 1) Frechet Inception Distance (FID) \cite{fid} between the 100 fake and 100 ground truth frames and 2) the average pixel distance between the synthesized and real images. As a feature extractor for FID, we employ the state-of-the-art identity recognition model, ArcFace \cite{Deng_2019_CVPR}.

\begin{figure}
\CenterFloatBoxes
\begin{floatrow}
\ffigbox
  {\includegraphics[width=\linewidth]{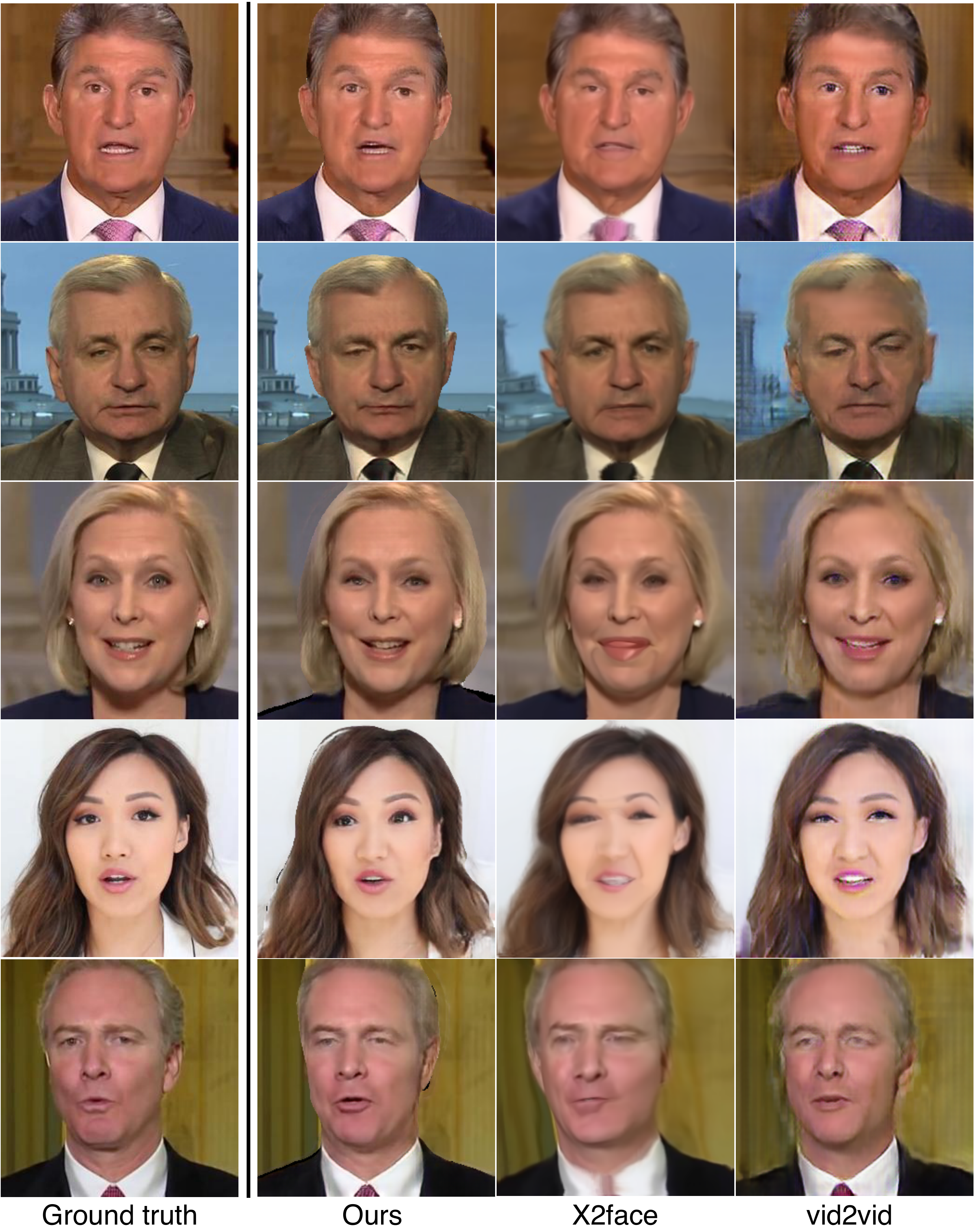}}
  {\caption{Comparison with X2face \cite{X2Face} and vid2vid \cite{vid2vid}.}
   \label{fig:fig3}}
\killfloatstyle
\ttabbox
  {
  \begin{tabular}{|c| c c c|}
    \hline
    \multicolumn{4}{|c|}{\textbf{Average pixel dist. ($\downarrow$)}}\\
    \hline
    & X2face & vid2vid & Ours \\
    \hline
    1 & 18.05 & 21.77 & \textbf{14.62} \\
    \hline
    2 & 15.48 & 19.11 & \textbf{13.09} \\
    \hline
    3 & \textbf{13.42} & 23.08 & 15.21 \\
    \hline
    4 & 26.54 & 35.23 & \textbf{25.74} \\
    \hline
    5 & 26.27 & 22.76 & \textbf{20.34} \\
    \hline
    \hline
    \multicolumn{4}{|c|}{\textbf{FID} ($\downarrow$)}\\
    \hline
    & X2face & vid2vid & Ours \\
    \hline
    1 & 0.32 $\pm$ 0.01 & 0.64 $\pm$ 0.01 & \textbf{0.23 $\pm$ 0.01} \\
    \hline
    2 & \textbf{0.26 $\pm$ 0.01} & 0.90 $\pm$ 0.02 & 0.36 $\pm$ 0.01 \\
    \hline
    3 & 0.50 $\pm$ 0.01 & 0.72 $\pm$ 0.01 & \textbf{0.35 $\pm$ 0.01} \\
    \hline
    4 & 1.09 $\pm$ 0.02 & 0.97 $\pm$ 0.02 & \textbf{0.70 $\pm$ 0.02} \\
    \hline
    5 & \textbf{0.47 $\pm$ 0.01} & 0.92 $\pm$ 0.02 & 0.67 $\pm$ 0.01 \\
    \hline
  \end{tabular}
  }
  {\caption{Numeric comparison with X2face and vid2vid (self-reenactment). In most cases, the average pixel distance and FID metrics confirm that our approach outperforms both baselines.
  }\label{tab:tab1}}
\end{floatrow}
\end{figure}

\textbf{Comparison with baselines.} We compare our method, both quantitatively and qualitatively with two strong baselines: 1) X2face \cite{X2Face}, a warping-based method and 2) vid2vid \cite{vid2vid}, a video synthesis method. For X2face we used the provided pre-trained parameters. For a fair comparison, we trained vid2vid on the same multi-person dataset with our model, on the NMFC-to-RGB video translation task. Then, given a new person, we fine-tuned vid2vid network on the same short training sequence we used to fine-tune our model, consisting of 50 frames. For all three methods, we performed the same self-reenactment experiment using the test split of each identity, for five identities. The results in Fig. \ref{fig:fig3} indicate that in all cases our method outperforms the two baselines, in terms of image quality and ground-truth reconstruction. This is confirmed by the FID and average pixel distance metrics, in Table \ref{tab:tab1}.

\begin{figure}[h!]
    \centering
    \includegraphics[width=0.98\linewidth]{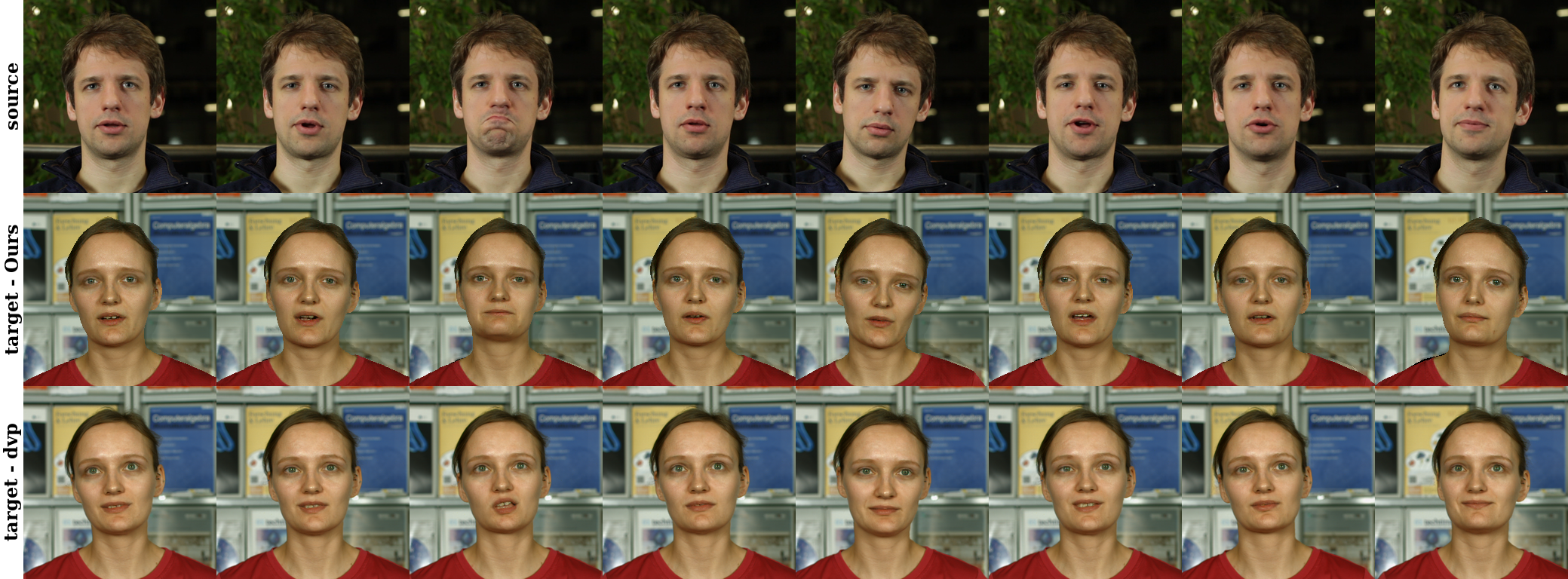}
    \caption{Comparison of our method with DVP \cite{deepvideoportraits}. 
    The source and generated sequences were kindly provided by the authors of DVP.}
    \label{fig:fig4}
\end{figure}

\textbf{Comparison with Deep Video Portraits (DVP)} \cite{deepvideoportraits}\textbf{.} We further compare our work with one more method, designed specifically to tackle the problem of head reenactment. Their network was trained on a long target sequence, containing a great variability of poses and expressions. For a fair comparison, we fine-tuned our model on the same training data. After that, we performed a head reenactment experiment. As can be seen in the frame examples of Fig. \ref{fig:fig4}, our approach performs equally well in terms of photo-realism and in some case surpasses the pose and expression transferability of DVP.

\begin{figure}
\CenterFloatBoxes
\begin{floatrow}
\ffigbox
  {\includegraphics[width=0.8\linewidth]{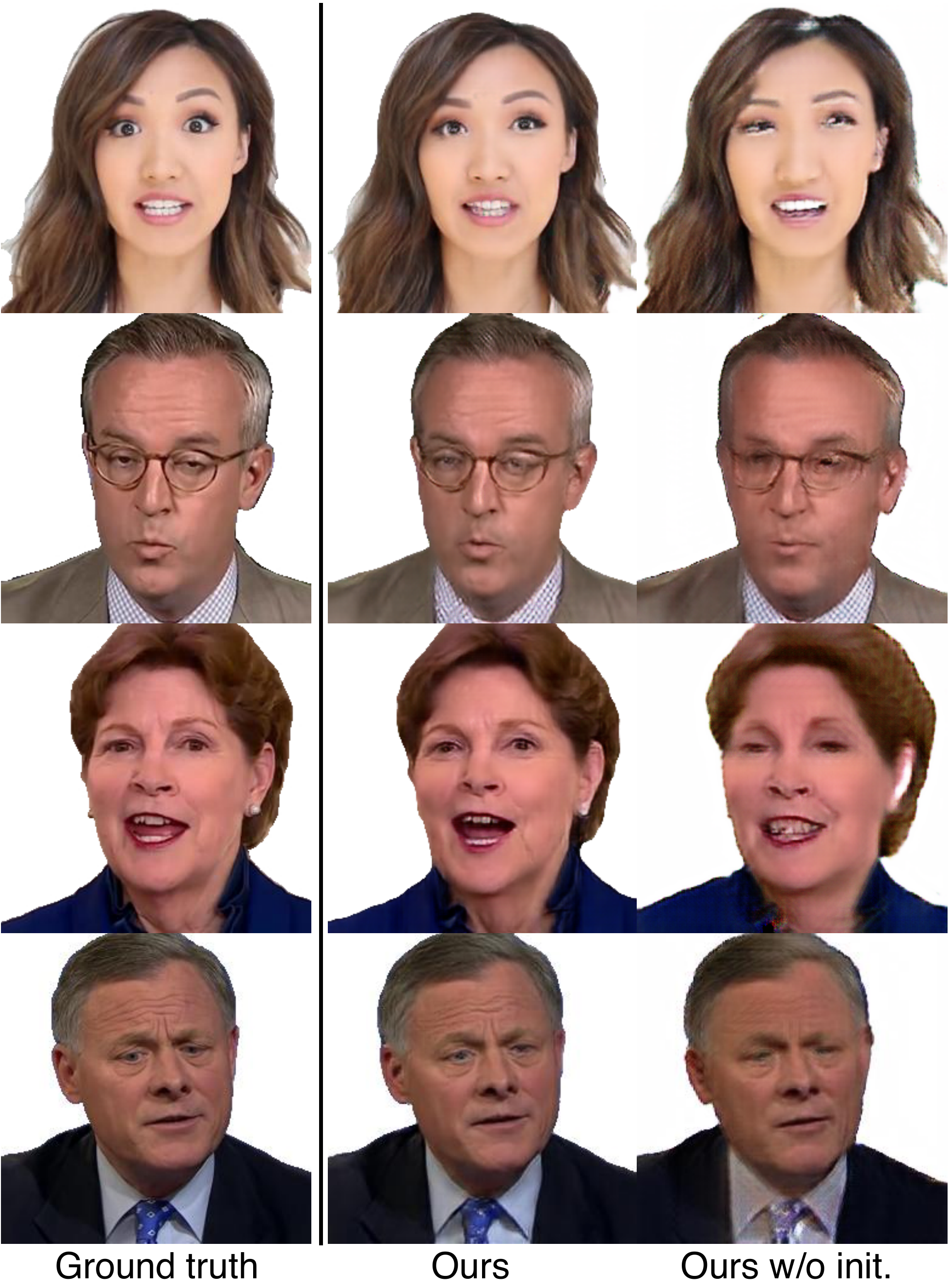}}
  {\caption{We demonstrate the significance of the initialization training stage using the multi-person dataset, in a self-reenactment scenario.}
  \label{fig:fig5}}
\killfloatstyle
\ttabbox
  {
  \footnotesize
  \begin{tabular}{|c| c c|}
    \hline
    \multicolumn{3}{|c|}{\textbf{Masked average pixel dist. ($\downarrow$)}}\\
    \hline
    & Ours & Ours w/o init.  \\
    \hline
    1 & \textbf{29.44} & 36.61   \\
    \hline
    2 &  \textbf{22.21} & 34.52 \\
    \hline
    3 &  \textbf{18.72} & 34.92 \\
    \hline
    4 &  \textbf{14.40} & 29.19 \\
    \hline
    \hline
    \multicolumn{3}{|c|}{\textbf{FID} ($\downarrow$)}\\
    \hline
    & Ours & Ours w/o init. \\
    \hline
    1 & \textbf{0.27 $\pm$ 0.01} & 0.67 $\pm$ 0.02 \\
    \hline
    2 & \textbf{0.16 $\pm$ 0.01} & 0.34 $\pm$ 0.02 \\
    \hline
    3 & \textbf{0.30 $\pm$ 0.01} & 0.65 $\pm$ 0.01 \\
    \hline
    4 & \textbf{0.20 $\pm$ 0.01} & 0.39 $\pm$ 0.01 \\
    \hline
  \end{tabular}
  }
  {\caption{Masked average pixel distance and FID scores under a self-reenactment experiment for four different identities, with and without the initialization training step. Results confirm the importance of training our framework on a multi-person dataset.}
  \label{tab:tab2}}
\end{floatrow}
\end{figure}

\textbf{Ablation study.} In order to demonstrate the importance of the model's initialization training stage, we performed a qualitative and quantitative study. We selected four identities, shown in Fig. \ref{fig:fig5}, and trained two variations of our model for each one. In the first case, we initialized the parameters of our framework with the values learned from the multi-person training stage, while in the second case we performed random initialization. We used 500 training frames for the identities appearing in the top two rows of Fig. \ref{fig:fig5}, and 250 frames for those shown in the two bottom rows. The importance of the initialization training stage is confirmed both visually and numerically in Table \ref{tab:tab2}, since FID scores and masked average pixel distances are significantly higher without the initialization step.

Next, we evaluate the effectiveness of the temporal discriminator $D^V$ and the dedicated mouth discriminator $D^M$ in the generative ability of our generative network. To that end, we trained our model and three more variations (without $D^M$, without $D^V$, without $D^M$ and $D^V$) on the multiple-person dataset for 16 epochs. Then, we computed the FID score and the average pixel distance for six random identities of the set, using their test data split for the self-reenactment experiment. The results are presented in Table \ref{tab:tab3}. This numeric evaluation indicates that removing the temporal discriminator has a large impact on the quality of the generated frames, while disconnecting the mouth discriminator deteriorates our results to a lesser extend, since the mouth is a small portion of the image.

\begin{table}[h!]
\scriptsize
\begin{center}
  \begin{tabular}{|c| c c c c | c c c c|}
    \hline
    \multirow{2}{*}{Id.} &
    \multicolumn{4}{c|}{\textbf{Average pixel dist. ($\downarrow)$}} &
    \multicolumn{4}{c|}{\textbf{FID} ($\downarrow$)} \\
    & Ours & w/o $D^M$ & w/o $D^V$ & w/o $D^V$, $D^M$ & Ours & w/o $D^M$ & w/o $D^V$ & w/o $D^V$, $D^M$ \\
    \hline
    \hline
    1 & 28.51 & 30.01 & 39.93 & 41.80 & 0.96 $\pm$ 0.01 & 1.55 $\pm$ 0.02 & 1.56 $\pm$ 0.01 & 1.79 $\pm$ 0.02 \\
    2 & 17.86 & 19.93 & 38.97 & 25.73 & 0.67 $\pm$ 0.01 & 1.39 $\pm$ 0.01 & 1.46 $\pm$ 0.02 & 1.70 $\pm$ 0.01 \\
    3 & 25.96 & 29.71 & 49.42 & 57.27 & 0.63 $\pm$ 0.01 & 1.20 $\pm$ 0.01 & 1.20 $\pm$ 0.01 & 1.44 $\pm$ 0.01 \\
    4 & 27.90 & 49.22 & 48.73 & 53.57 & 0.57 $\pm$ 0.01 & 1.14 $\pm$ 0.02 & 1.24 $\pm$ 0.01 & 1.45 $\pm$ 0.01 \\
    5 & 25.11 & 34.61 & 41.48 & 43.31 & 0.76 $\pm$ 0.02 & 1.62 $\pm$ 0.01 & 1.49 $\pm$ 0.02 & 1.83 $\pm$ 0.01 \\
    6 & 29.05 & 31.30 & 39.34 & 45.96 & 0.82 $\pm$ 0.01 & 1.35 $\pm$ 0.02 & 1.49 $\pm$ 0.03 & 1.63 $\pm$ 0.02 \\
    \hline
  \end{tabular}
\end{center}
   \caption{Numeric evaluation of the importance of $D^M$ and $D^V$.}
  \label{tab:tab3}
\end{table}

\begin{figure}[h!]
    \centering
    \includegraphics[width=0.95\linewidth]{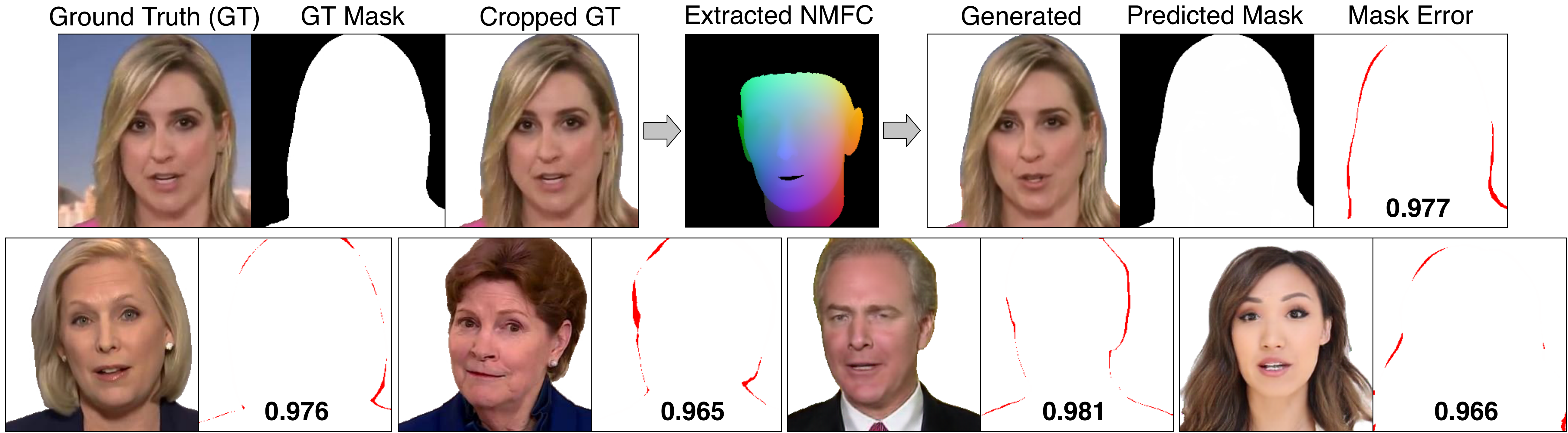}
    \caption{Average IoU across test frames for 5 identities, reported at the lower part of the images that show the mask error between the predicted and GT mask.}
    \label{fig:figmasks}
\end{figure}

In order to validate the predictive performance of our generator on the foreground mask, we conducted one more experiment: We fine-tuned five models on the identities shown in Fig. \ref{fig:figmasks} above and computed the average Intersection over Union (IoU), showing that the predicted foreground masks are very accurate.


\textbf{Automated Study.} Rossler et al. \cite{roessler2019faceforensicspp} proposed a method for detecting fake videos created by facial manipulation methods. Their classifier was trained on a large dataset, created by manipulating $1,000$ YouTube videos, with graphics-based \cite{face2face}, \cite{Marek2019} and learning-based \cite{torzdf2019}, \cite{DeferredNeuralRendering}, facial reenactment methods. They reported a high detection accuracy, around $99 \%$ on raw video data. We utilise this network to assess the realism of our synthesised videos automatically. More specifically, we randomly selected 10 synthetic videos produced with our reenactment method and run them through the manipulation-detection network of \cite{roessler2019faceforensicspp}, obtaining an accuracy of $25\%$. This demonstrates that distinguishing our synthetic videos from real ones is challenging even for such a well-trained system.

\section{Conclusion}

We proposed \textit{head2headFS}, a pipeline consisting of a 3D facial reconstruction system and a video rendering network. Our NMFC representation enables performing full head reenactment from a source to a target without distorting the identity of the latter. Our few-shot based training strategy allows to fine-tune our model on a new identity very quickly. Most importantly, during test our entire pipeline works in real-time fashion. 

\bibliographystyle{splncs04}
\bibliography{eccv2020submission}

\end{document}